\documentclass[10pt,twocolumn,letterpaper]{article}

\usepackage[pagenumbers]{cvpr} 

\definecolor{cvprblue}{rgb}{0.21,0.49,0.74}
\usepackage[pagebackref,breaklinks,colorlinks,allcolors=cvprblue]{hyperref}
\usepackage{multirow}
\usepackage{multicol}
\usepackage{amssymb}
\usepackage{colortbl}
\usepackage{bm} 
\usepackage{pifont}

\definecolor{darkgreen}{RGB}{34,139,34}

\def\paperID{*****} 
\def\confName{CVPR}
\def\confYear{2026}

\title{S2D: Sparse-To-Dense Keymask Distillation for \\ Unsupervised Video Instance Segmentation}

\author{Leon Sick\\
Ulm University\\
\and
Lukas Hoyer\\
Google\\
\and
Dominik Engel\\
KAUST\\
\and
Pedro Hermosilla\\
TU Vienna\\
\and
Timo Ropinski\\
Ulm University\\
}

\begin{document}

\twocolumn[{
\maketitle
\vspace{-1.5em} 
\begin{center}
  \centering
    \includegraphics[width=\textwidth, page=1]{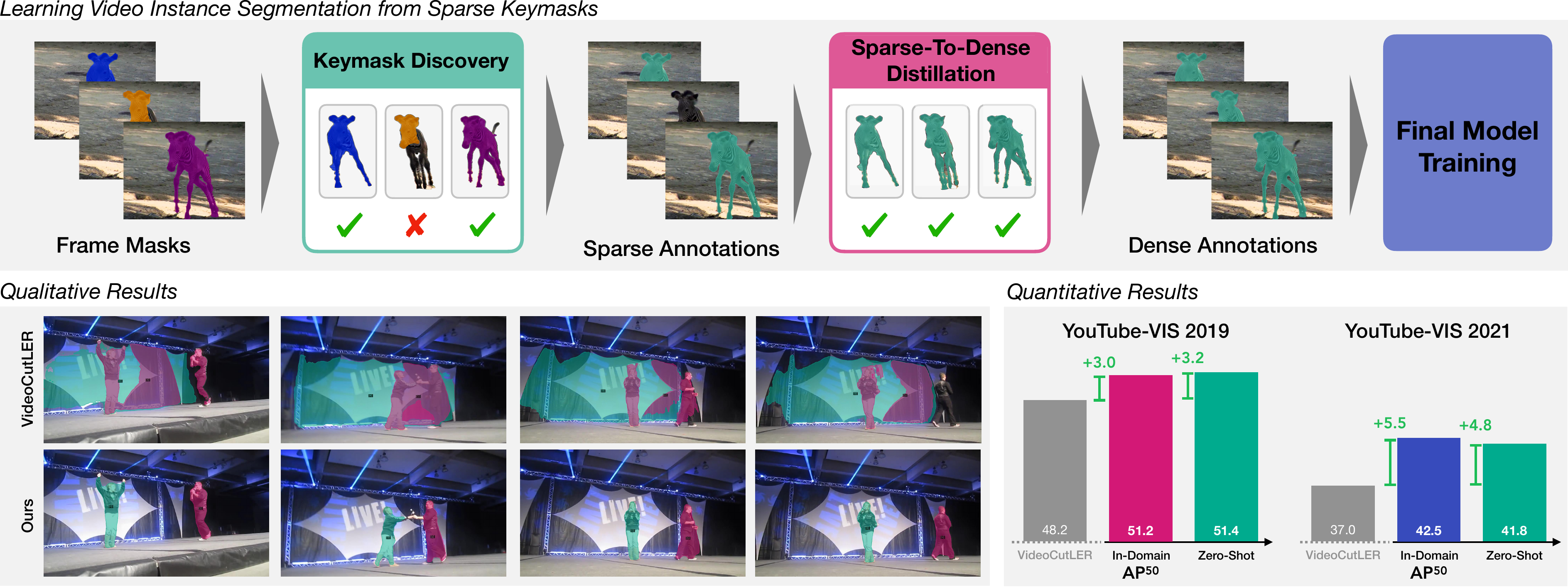}
    \captionof{figure}{\textbf{Learning Unsupervised Video Instance Segmentation from Real Videos.} Given single-frame unsupervised instance masks, we first discover temporally-coherent, high-quality keymasks. This sparse labelset is then propagated using Sparse-To-Dense Distillation, aided by Temporal DropLoss. Finally, we train our model on the resulting dense labelset and demonstrate state-of-the-art results.}
    \label{fig:main}
\end{center}
}]

\begin{abstract}
In recent years, the state-of-the-art in unsupervised video instance segmentation has heavily relied on synthetic video data, generated from object-centric image datasets such as ImageNet. However, video synthesis by artificially shifting and scaling image instance masks fails to accurately model realistic motion in videos, such as perspective changes, movement by parts of one or multiple instances, or camera motion. To tackle this issue, we propose an unsupervised video instance segmentation model trained exclusively on real video data. We start from unsupervised instance segmentation masks on individual video frames. However, these single-frame segmentations exhibit temporal noise and their quality varies through the video. Therefore, we establish temporal coherence by identifying high-quality keymasks in the video by leveraging deep motion priors. The sparse keymask pseudo-annotations are then used to train a segmentation model for implicit mask propagation, for which we propose a Sparse-To-Dense Distillation approach aided by a Temporal DropLoss. After training the final model on the resulting dense labelset, our approach outperforms the current state-of-the-art across various benchmarks. Project Page: \href{https://leonsick.github.io/s2d/}{leonsick.github.io/s2d}
\end{abstract}    

\section{Introduction}
\label{sec:intro}
Video instance segmentation is a key task in perceptive computer vision, empowering a range of applications, such as autonomous driving, AR/VR applications, and video editing. Compared to image instance segmentation, video instance segmentation though is fundamentally more challenging. To segment videos, models must learn temporal dynamics and maintain identity-preserving masks, while instances may appear, disappear and undergo occlusions. In recent years, a range of supervised Video Instance Segmentation models have advanced rapidly~\cite{yang2019video,cheng2021mask2formervideoinstancesegmentation,wang2021end,wu2022seqformer,zhang2023dvis}, steadily increasing segmentation accuracy. However, many of these models rely on per-frame human annotations, which require immense annotation efforts and thus result in high costs. For example, the largest video segmentation dataset yet, SA-V~\cite{ravi2024sam2}, contains 190K manual annotations, for which the authors employed a large group of crowdworkers. Such costly efforts have sparked the field of Unsupervised Video Instance Segmentation. This research area follows the same core motivation as unsupervised image segmentation: Enabling segmentation without task-specific or any other human annotations, therefore removing entirely the need for human labeling anywhere in the pipeline \cite{wang2023cut, sick2025cuts3d, arica2024cuvler, wang2024videocutler, stego, sick2024depthg, Hahn_2025_cups}.
Wang et al.~\cite{wang2024videocutler} have proposed VideoCutLER, a video segmenter trained on synthesized sequences of shifting instances through object-centric images from ImageNet~\cite{deng2009imagenet}. Critically, their training data models only single-object translational movement examples, that lack the complexity of multiple dynamic objects in real-world videos.
To overcome this limitation, we enable our model to learn complex temporal dynamics, involving multi-instance motion, from real-world videos without requiring human annotations.

Specifically, we propose \textbf{S2D}, an approach for training a video instance segmentation model to make dense predictions by training on a sparse set of temporally-coherent, high-quality keymasks, predicted by an unsupervised image instance segmenter.
We achieve this relying on three key insights. First, we can extract high-confidence masks for certain frames with existing image unsupervised instance segmentation models. Second, we can identify which frames contain reliable masks by inspecting their temporal coherence. Lastly, we can train a video instance segmentation model on sparse image annotations via Sparse-To-Dense distillation to achieve mask propagation.
Our proposed pipeline is illustrated in Figure~\ref{fig:main}.
In the first step, our Keymask Discovery algorithm identifies the quality of frame-wise image masks without human supervision by leveraging point tracks as a deep motion prior. The resulting masks are identified as keymasks of the video, i.e., temporally-coherent instance masks.
We predict single-frame masks using an off-the-shelf unsupervised image instance segmentation model. By tracking each single-frame mask over the entire video sequence, we identify when appearances and disappearances occur. Performing this analysis for all single-frame masks yields visibility groupings, which contain instance masks whose proxy-propagations, i.e., point tracks, appear and disappear together.
Following, we determine which of the simultaneously visible masks correspond to each other by applying our temporal correspondence matching. Using the overlap of proxy-propagations and per-frame instance masks, we subdivide the visibility groups into clusters that share temporal and spatial correspondence.
This process yields sparse video annotations, i.e., the video instances have annotations for some frames where high-quality masks could be identified.
However, our method aims to obtain mask annotations for video instances in every frame where they are visible, i.e., a dense annotation set.
To achieve this, we demonstrate that a VideoMask2Former~\cite{cheng2021mask2formervideoinstancesegmentation} can be trained as an implicit mask propagation \& object discovery model.
We propose a novel approach for training this model on sparse annotations to complete the missing video instance masks. First, we waive the loss penalty for missing instance annotations by proposing a Temporal DropLoss. Second, we propose Sparse-To-Dense Distillation, where we train a student model to learn dense predictions, guided by a teacher model.
Sparse-To-Dense Distillation results in a dense video annotation set, which we further refine in another round of training.
Since our process to generate pseudo-labelsets is applicable to any video dataset, we experiment with scaling the training video data beyond the in-domain data and find that this yields strong zero-shot performance.

In summary, we propose the following contributions:
\begin{enumerate}[label=\arabic*.]
    \item We propose a \textbf{Keymask Discovery} algorithm to construct a sparse pseudo-labelset of high-quality temporally-coherent instance masks.
    
    \item We introduce \textbf{Sparse-To-Dense Distillation} with \textbf{Temporal DropLoss} which achieves implicit mask propagation \& object discovery. This allows our method to convert the sparse keymasks to a dense video labelset.

    \item We present a video instance segmentation model, trained without task-specific labels and exclusively on real videos, which demonstrates state-of-the-art \textbf{in-domain and zero-shot} performance.

\end{enumerate}

\section{Related Work}
\label{sec:related-work}

\noindent\textbf{Unsupervised Image Instance Segmentation.} 
Early methods for unsupervised instance segmentation have been focused on leveraging features from self-supervised learning models to extract masks. Taking a bottom-up approach, MaskDistill~\cite{van2022maskdistill} leverages a pixel grouping prior from MoCo features to extract masks, which are then used to train a standard detection network like Mask R-CNN~\cite{he2017mask}. In contrast, FreeSOLO~\cite{wang2022freesolo} leverages DenseCL features to extract attention maps based on a key-query mechanism. The resulting maps are used as pseudo-masks to train a SOLO model~\cite{wang2021solo}.

Another significant direction extends Normalized Cut-based segmentation techniques. Wang et al. propose CutLER~\cite{wang2023cut} by generalizing the use of Normalized Cut on DINO~\cite{caron2021emerging} features for mask extraction, first introduced by TokenCut~\cite{wang2023tokencut}, to the multi-instance case. After extracting pseudo-masks with MaskCut, a detection network is trained for multiple rounds with self-training. CuVLER, by Arica et al.~\cite{arica2024cuvler}, enhances the pseudo-mask generation by using a six-DINO-model ensemble and a soft target loss.
Finally, CutS3D~\cite{sick2025cuts3d} leverages 3D information to separate instances from semantics. LocalCut is used on a 3D point cloud to cut the instances, and the detector's training is augmented with Spatial Confidence components. The resulting model achieves improved instance separation compared to previous state-of-the-art methods for zero-shot and in-domain evaluations. We utilize CutS3D in our work to predict single-frame instance masks.
\\

\noindent\textbf{Unsupervised Video Instance Segmentation.}
Unsupervised video instance segmentation is a challenging task that requires not only the separation of foreground objects from the background but also the differentiation and tracking of multiple distinct instances over time, all without any human annotations \cite{zhou2022survey}. 
Previous research on unsupervised video segmentation largely proposed approaches for unsupervised video object segmentation ~\cite{karazija2024learning, xie2022segmenting, xie2024appearance, safadoust2023multi}. VOS aims to detect all moving objects as a single foreground, without the necessity of distinguishing between individual object instances. A common practice in prior studies on this task \cite{ventura2019rvos, xie2022segmenting, xie2024appearance, yang2021motiongroup} is the dependence on optical flow networks.
Notably, work by Karazija et al.~\cite{karazija2024learning} has leveraged model-predicted point tracking and optical flow as supervision for their segmentation model. MotionGroup~\cite{yang2021motiongroup} train a transformer to extract motion-grouped segmentations using only optical flow and no RGB frames. Furthermore, OCLR~\cite{xie2022segmenting} propose an object-centric approach, also leveraging model-predicted optical flow.
Another line of work tackles unsupervised label propagation throughout a video sequence, using human annotations only for the first frame \cite{caron2021emerging, jabri2020crw, tang2023emergent} or by incorporating supervised learning with extensive external labeled data \cite{luiten2020unovost}. For unsupervised video instance segmentation, VideoCutLER~\cite{wang2024videocutler} has proposed  copy-paste shifting ImageNet~\cite{deng2009imagenet} masks to generate synthetic videos. The scale of their synthetic dataset allows the method to scale to a zero-shot model.

\section{Method}
\label{sec:method}
\begin{figure}[!th]
  \centering
    \includegraphics[width=\linewidth, page=2]{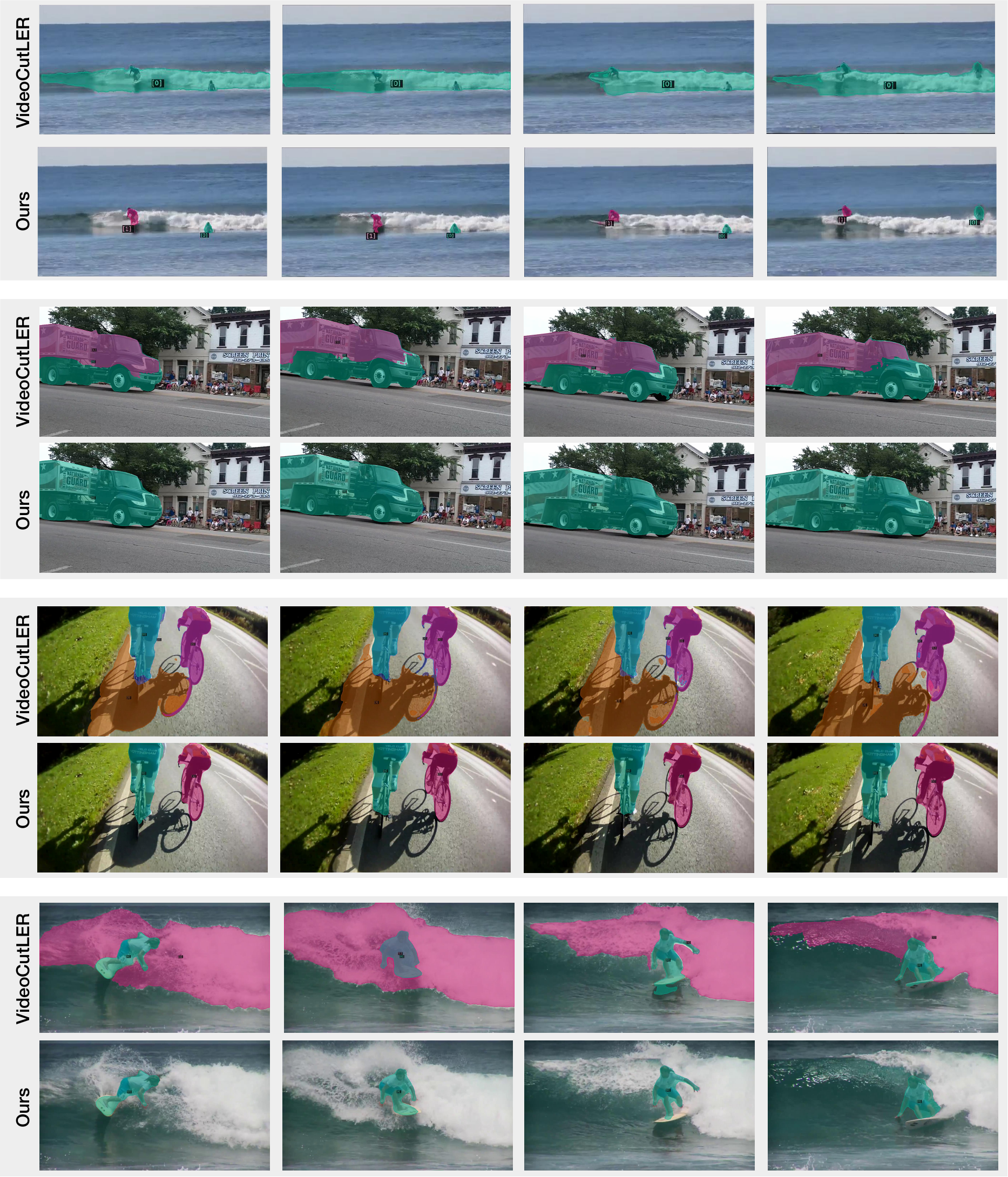}
    \caption{\textbf{Qualitative Results.} Our model is able to segment more fine-grained structures in videos compared to VideoCutLER~\cite{wang2024videocutler}. Furthermore, we find our predictions are less noisy.} 
    \label{fig:qualitative}
    \vspace{-4mm}
\end{figure}
\begin{figure*}[!ht]
  \centering
    \includegraphics[width=\textwidth, page=3]{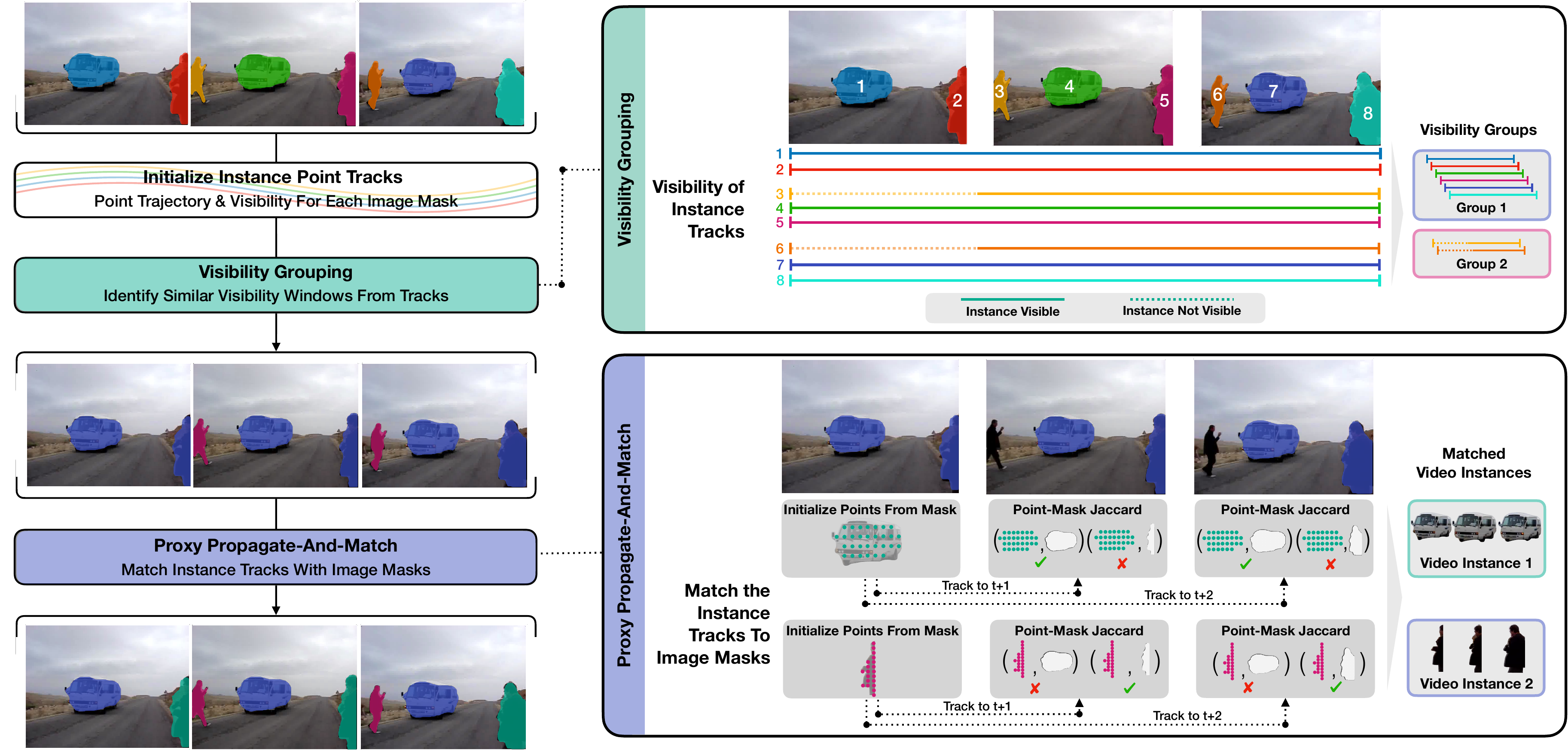}
    \caption{\textbf{Keymask Discovery.} Given single-frame unsupervised instance masks, we discover temporally-coherent keymasks by first generating instance tracks, then matching image instance masks based on temporal coherence concepts.\\
    \textbf{Visibility Grouping:} Instance Tracks of Single-Frame Masks belonging to the same video instance have similar visibility windows.\\
    \textbf{Proxy Propagate-And-Match:} In a Visibility Group, frame masks can be temporally associated if their instance tracks match other masks.} 
    \label{fig:keymask}
    \vspace{-3mm}
\end{figure*}

An overview of our method is presented in \cref{fig:main}: Starting from noisy single-frame masks, we perform Keymask Discovery to obtain a sparse pseudo-annotation set. Next, we propagate the extracted temporally-coherent masks using Sparse-To-Dense Distillation, resulting into a dense annotation set. Finally, we train the final model on the dense pseudo-labels.
Specifically, we use CutS3D~\cite{sick2025cuts3d} to first extract frame-wise instance masks for all videos in the dataset. We then filter those predictions to identify sparse, temporally-coherent keymasks (see Figure~\ref{fig:keymask} and \cref{sec:keymasks}), which are then used to train our implicit mask propagation module using our proposed Temporal DropLoss (\cref{sec:droploss}). To train this model, we employ a Sparse-To-Dense Distillation setup, which allows the model to smooth out ambiguities remaining in the pseudo-annotations in a dual-stage setup (Figure~\ref{fig:sparse2dense} and \cref{sec:dual-stage}). The resulting model predicts high-quality video instance masks, as is shown in \cref{fig:qualitative}.

\subsection{Temporal Grouping for Keymask Discovery}
\label{sec:keymasks}
We start with unsupervised single-frame predictions from CutS3D~\cite{sick2025cuts3d} as the basis for video pseudo annotation. However, in a video context, the CutS3D predictions lack any classification or temporal coherence, i.e., they do not encode which mask in frame $1$ belongs to which mask in frame $2$. Furthermore, there might be inconsistencies in a video across single-frame predictions. 
To establish temporal coherence and filter out noisy masks, we propose a novel unsupervised keymask discovery method. 
\\
\noindent\textbf{Obtaining Mask-Point Trajectories.} 
Using an off-the-shelf point tracker~\cite{karaev2025cotracker3}, we initialize a sparse point grid for each single frame mask and track it forward/backward through the video. 
Practically, we initialize a set of $N_i$ instance points $\mathbf{P}_i^t = \{p_1^t, \dots, p_{N_i}^t\}$ for a given instance mask $m_i^t$ of instance $i$ at time $t$. We track all points throughout the video for all time steps $t \in \{1, \dots, T\}$, i.e. forward \& backward, and record their coordinates and visibilities. This results in a set of trajectories $\mathcal{T}_i = \{\tau_1 , \dots, \tau_{N_i} \}$ for each instance $i$, where each trajectory $\tau_j = \{(x_j^t, v_j^t) \mid t= 1,\dots,T\}$ encapsulates the coordinates $x_j^t \in \mathbb{R}^2$ and visibility $v_j^t \in \{0,1\}$ of point $j$ over the video duration. 
With trajectory and visibility data present for each single-frame mask in the video, we employ it as a proxy to validate our fundamental temporal coherence concepts.
\\
\noindent\textbf{Visibility Grouping.} 
In the next step, we validate that single-frame mask tracks of the same temporal instance appear and disappear at the same time, and we verify this concept in the following section. 
For each single-frame instance mask $m_i^t$, we determine if its proxy-propagation, given by the instance trajectories $\mathcal{T}_i$, is visible at any given timestep $t$ in the video. This is the case if the proportion of tracked points that are visible at a given frame, $\gamma_t = \frac{1}{N_i} \sum_{j=1}^{N_i} v_j^t$, exceeds a threshold $\gamma_\text{thr}$.
In practice, we consider an instance as occluded if the ratio of visible points is below $\gamma_{\text{thr}}=0.3$. 
We collect these binary visibility tracks, displayed in \cref{fig:keymask}, for each single-frame mask across the video and cluster them with DBSCAN~\cite{dbscan}. The input to the clustering algorithm is the set of binary visibility vectors $\mathcal{V} = \{ \mathbf{v}_1 , \dots, \mathbf{v}_M \}$, where each vector $\mathbf{v}_i = \{ \mathbf{1}(\gamma_t > \gamma_\text{thr}) \mid t = 1,\dots,T\}$
represents the sequence of visibility states for a single mask $m_i^t$ across the video duration and $\mathbf{1}(\cdot)$ is the indicator function that is $1$ if the condition is true and $0$ otherwise.
The resulting clusters $\mathcal{C} \subset \mathcal{V}$ contain all single-frame masks whose proxy-propagations appear and disappear together, i.e., the instance tracks are visible at the same time. We find this provides a useful initial grouping to find masks of temporally-coherent instances. In the example in \cref{fig:keymask}, the human on the left appears later, hence it is a different video instance. In addition, the clustering returns outlier detections $\mathcal{O} \subset \mathcal{V}$ for masks that cannot be grouped by visibility. These are often noisy masks which do not belong to any instance. 
\\
\noindent\textbf{Proxy Propagate-And-Match.}
For cases where each video instance has separate visibility windows, our Visibility Grouping identifies video instances correctly. However, there are many cases where multiple instances have the same visibility, e.g., both are visible throughout the entire video, as in the case for the van and the person on the right in \cref{fig:keymask}. Hence, we further subdivide the visibility groups using our proxy propagate-and-match approach.
We validate our second temporal coherence concept by measuring the spatial consistency between instance masks across frames. Specifically, the point tracks associated with a single-frame instance mask are used as a proxy to propagate that instance forward and backward to a different frame. This proxy-propagation must exhibit high overlap with the single-frame mask in that target frame. We restrict this search to masks within the same Visibility Group, which serves as an effective initial grouping. \\
To compute this overlap, we again leverage the instance trajectories $\mathcal{T}_i$ for each single-frame mask within a visibility cluster $\mathcal{C}$. For a reference instance $i$ and its propagated point set $\mathbf{X}_i^t = \{x_1^t , \dots, x_{N_i}^t\}$ at the frame of timestep $t$, we calculate the degree of overlap with every instance mask $m_k^t$ in this frame.
Specifically, we propose to compute the point-mask intersection between the point set $\mathbf{X}_i^t$ and the instance masks $m_k^t$ using the Point-Mask Jaccard Index $\mathcal{J}(i, k, t)$, which measures the overlap of the propagated points of instance $i$ and the mask of image instance $k$ of the frame at time $t$.
Given that $\mathbf{X}_i^t$ is a sparse set of $N_i$ points and $m_k^t$ is a dense set of pixels, the Point-Mask Jaccard Index is formally expressed as:
\begin{equation}
    \mathcal{J}(i, k, t) = \frac{\sum_{j=1}^{N_i} \mathbf{1}(x_j^t \in m_k^t)}{N_i} , \quad x_j^t \in \mathbf{X}_i^t
\end{equation}
If $\mathcal{J}(i, k, t) > \lambda_{\mathcal{J}}$, we consider the point grid $\mathbf{X}_i^t$ matched with the mask $m_k^t$. In practice, we set $\lambda_{\mathcal{J}} = 0.5$.

We compute the matches for all instance point-tracks to all instance masks within a visibility cluster $\mathcal{C}$. The matching process yields a binary matching matrix $\mathbf{M} \in \{0, 1\}^{N_{\text{masks}} \times N_{\text{masks}} \times (T)}$, where the match is defined by the following indicator function:
\begin{equation}
\mathbf{M}_{i, k}^t = \mathbf{1}(\mathcal{J}(i, k, t) > \lambda_{\mathcal{J}})
\end{equation}
We collect the set of all matching tracks where each element represents the temporal sequence of matches for a point track $i$ . We then cluster the resulting matching tracks again with DBSCAN~\cite{dbscan}, which provides matched subgroups within a visibility cluster. In the case of the van-person visibility group in \cref{fig:keymask}, this leads to both instances being separated successfully.
With this, we arrive at the final instance mask assignments over time, which meet the defined temporal coherence concepts. Since all masks which do not meet these criteria are filtered out, this pseudo-annotation set is inherently sparse. For cases where no temporal coherence can be established from the image instance masks, the video is discarded from the annotation set.
Note that other unsupervised video segmentation methods have also relied on model-predicted motion cues~\cite{karazija2024learning, luiten2020unovost, yang2021motiongroup, xie2022segmenting, xie2024appearance}. However, contrary to existing approaches, our method does not use point tracks or optical flow as supervision to the model.

\subsection{Temporal DropLoss For Mask Propagation}
\label{sec:droploss}
Our goal is to obtain a model that makes temporally dense predictions by learning only from sparse annotations, thereby achieving implicit mask propagation. From the Keymask Discovery process, we obtain a sparsely annotated pseudo-label set, i.e., not every frame might contain a segmentation mask, even though the instance is present.
In order to compensate for this training data sparsity, we make several modifications to our training pipeline. First, we modify the dataloading process: Instead of sampling frames randomly throughout the entire video, we prioritize frames with at least one temporally dense annotated instance mask, thereby constructing pseudo-dense training samples. Training on this data gives the model some samples with annotated temporal consistency. However, these sampled video snippets do not only contain dense mask annotations, but also sparsely labeled instances. For example, a second instance in the video snippet might only have a mask for 1 of 2 frames. Computing a standard video instance segmentation loss on this sample would assign a penalty for predicting the second, un-annotated mask.
To address this, we propose a Temporal DropLoss for video instance segmentation: For instances with sparse annotations, we limit the loss computation to the frame(s) where a given instance is annotated, and drop the loss for frames without an annotated mask. We find this encourages mask propagation to sparsely labeled instances since the model has learned to propagate temporally-coherent masks from the densely annotated instance samples in the data.
In practice, we use a VideoMask2Former~\cite{cheng2021mask2formervideoinstancesegmentation} as our video instance segmentation model, which employs bipartite matching between pseudo ground-truth masks and its predictions, to then compute the mask losses only for these matches. Consequently, we formulate our Temporal DropLoss for the matched predictions on the mask loss of the model. 
Given a set of sparse instance mask annotations $\mathbf{A}^t$ and matched model predictions $\hat{\mathbf{M}}^t$, we drop the segmentation loss $\mathcal{L}_{\text{mask}}$ for a given instance $i$ at frame $t$ if the sum of its pseudo-annotated binary mask $\mathbf{m}_{i}^t$ is zero, i.e. no matching mask is present at $t$. For a matched instance prediction, the loss is formally defined as:
\begin{equation}
\mathcal{L}_{\text{TempDrop}}(i) = \sum_{t=1}^{T} \mathbf{1}(\|\mathbf{m}_{i}^t\| > 0) \mathcal{L}_{\text{mask}}(\hat{\mathbf{m}}_{i}^t, \mathbf{m}_{i}^t)
\end{equation}
where $\|\mathbf{m}_{i}^t\|$ denotes the total number of pixels in the pseudo ground-truth mask.
$\mathcal{L}_{\text{mask}}$ is the standard Mask2Former~\cite{cheng2022mask2former} mask loss which combines a binary cross-entropy loss with a dice loss~\cite{milletari2016dice}, i.e. 
\begin{equation}
\mathcal{L}_{\text{mask}} = \lambda_{\text{ce}}\mathcal{L}_{\text{ce}}+\lambda_{\text{dice}}\mathcal{L}_{\text{dice}}
\end{equation}
Our approach can be understood as a the temporal extension of Image DropLoss~\cite{wang2023cut}, which is a popular technique in unsupervised image instance segmentation to counteract the penalty of predicting masks that are completely disjoint from the pseudo-annotations.

\begin{figure}[!t]
  \centering
    \includegraphics[width=\linewidth, page=4]{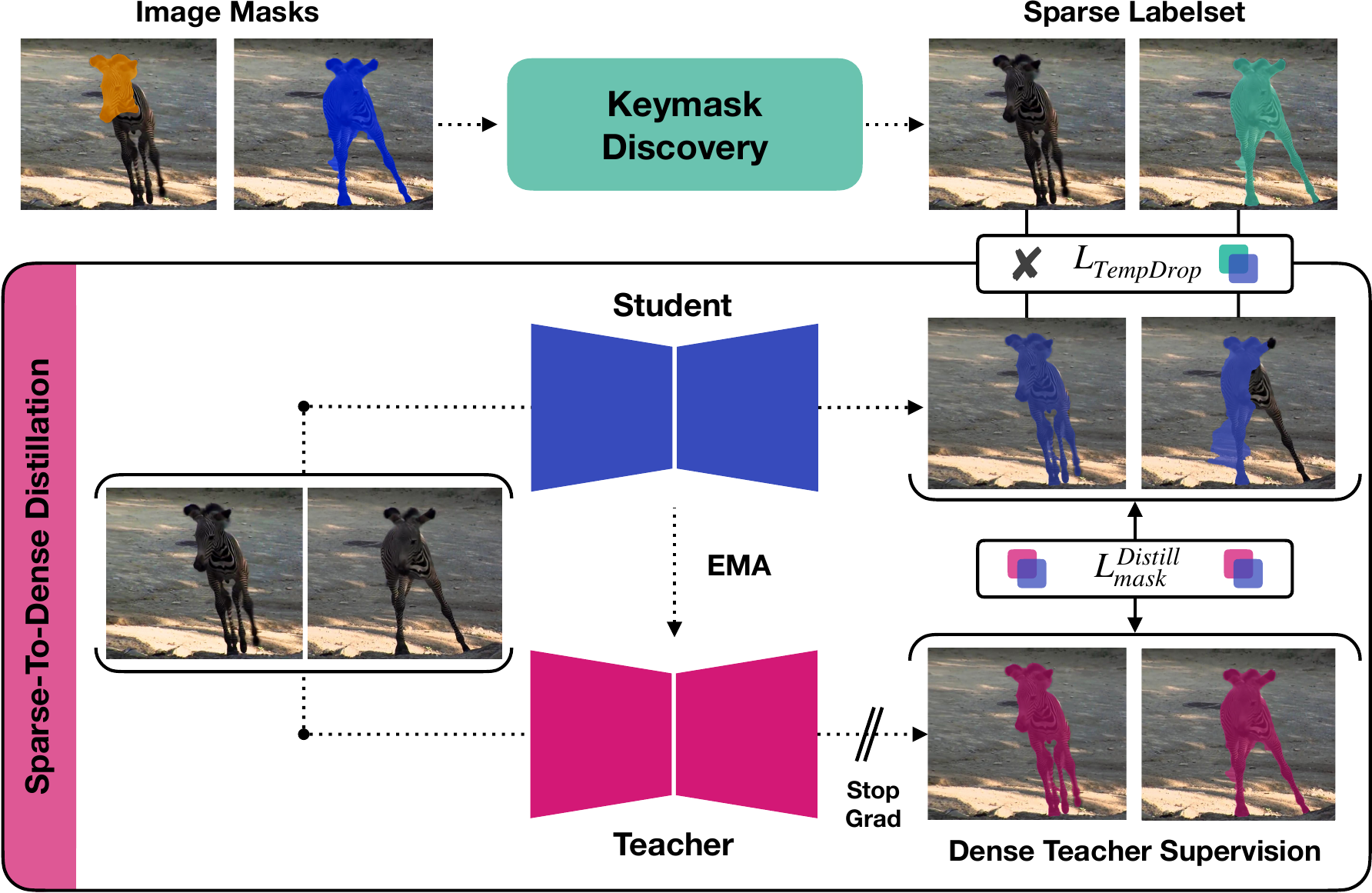}
    \caption{\textbf{Sparse-To-Dense Distillation.} We supervise the student model with our Temporal DropLoss on the sparse anchor labelset. Additionally, the teacher model provides dense supervision on-the-fly, which improves as training progresses.} 
    \label{fig:sparse2dense}
\end{figure}

\subsection{Dual-Stage Distillation Training}
\label{sec:dual-stage}
To train our final video instance segmentation model, we use a dual-stage training setup: In the first stage, we train a VideoMask2Former on our sparse keymask annotations to achieve implicit mask propagation throughout the training dataset. In the second stage, we train the final video instance segmentation model on the predicted dense annotations.
In the first step, we propose a Sparse-To-Dense distillation approach for training a video instance segmentation model on temporally sparse annotations. We set up a student-teacher distillation with both models being a VideoMask2Former. The student receives two learning signals. We train the student model on the sparse keymask annotations with the Temporal DropLoss, where the keymasks serve as an anchoring labelset. The loss already encourages the student to learn temporally-consistent implicit mask propagation. 
We update the teacher model weights with an exponential moving average (EMA), controlled by the update rate $\mu$. This way, the teacher becomes better at predicting temporally-dense video instance segmentations as the training progresses. Harvesting this signal, we task the student model to match the dense teacher predictions. This enables the student to refine its temporal coherence capabilities by learning temporally-consistent predictions from the teacher on-the-fly. In addition, maintaining supervision from the sparse keymasks prevents the model from deviating too far from the pseudo ground-truth. With this, the full loss can be formulated as
\begin{equation}
\mathcal{L}_{\text{Full}} =\mathcal{L}_{\text{TempDrop}}+\mathcal{L}_{\text{mask}}^\text{Distill}
\end{equation}
We illustrate our Sparse-To-Dense Distillation in \cref{fig:sparse2dense}. 
In the second stage, we aim to further refine the model's predictions. We employ the same student-teacher setup again. Using the final Sparse-To-Dense model from stage 1, we predict a new set of now dense pseudo-annotations to serve as anchoring labelset. The new student model learns from these temporally dense pseudo-annotations and is additionally tasked to match the teacher's predictions. For the second round of distillation, the model learns refined instance segmentations by self-distilling its own predictions. We also observe that keeping the dense anchoring annotation set as supervision helps the model to converge.

\begin{table*}[!th]
    \centering
    \caption{\textbf{Unsupervised Video Instance Segmentation.} When trained on in-domain data, our model is able to outperform the previous state-of-the-art across YouTubeVIS-2019, YouTubeVIS-2021 and DAVIS-all. \textsuperscript{*} Results obtained using official checkpoint.}
    \vspace{-2mm}
    \resizebox{\linewidth}{!}{
    \begin{tabular}{lccccccccccccccc}
        \toprule
        & \multicolumn{6}{c}{YouTube-VIS 2019} & \multicolumn{6}{c}{YouTube-VIS 2021} & \multicolumn{3}{c}{DAVIS-all} \\
        \cmidrule(lr){2-7} \cmidrule(lr){8-13} \cmidrule(lr){14-16} 
        Method & AP$_{50}$ & AP$_{75}$ & AP & AP$_{\text{S}}$ & AP$_{\text{M}}$ & AP$_{\text{L}}$ & AP$_{50}$ & AP$_{75}$ & AP & AP$_{\text{S}}$ & AP$_{\text{M}}$ & AP$_{\text{L}}$ & $J\&F$ & $J$-Mean & $F$-Mean\\
        \midrule
        MotionGroup~\cite{yang2021motiongroup} & 0.5 & 0.0 & 0.1 & 0.0 & 0.4 & 0.1 & 0.3 & 0.0 & 0.1 & 0.0 & 0.2 & 0.1 & - & - & -  \\
        OCLR~\cite{xie2022segmenting} & 5.5 & 0.3 & 1.6 & 0.1 & 1.6 & 6.1 & 4.3 & 0.1 & 0.9 & 0.1 & 1.0 & 4.9 & 36.9 & 36.3 & 37.6  \\
        CutLER~\cite{wang2023cut} & 36.4 & 13.5 & 16.0 & 3.5 & 13.9 & 26.0 & 28.5 & 9.3 & 12.0 & 2.7 & 12.7 & 27.2 & 44.7 & 44.6 & 44.9  \\
        VideoCutLER\textsuperscript{*}~\cite{wang2024videocutler} & 48.2 & 22.8 & 24.5 & \textbf{7.0} & 17.5 & 36.1 & 37.0 & 16.1 & 18.0 & 4.9 & 17.0 & 36.8 & 44.9 & 44.8 & 45.0 \\
        S2D (Ours) & \textbf{51.2} & \textbf{25.2} & \textbf{26.2} & 6.0 & \textbf{21.2} & \textbf{39.4} & \textbf{42.5} & \textbf{17.6} & \textbf{20.1} & \textbf{5.1} & \textbf{21.7} & \textbf{41.0} & \textbf{50.6} & \textbf{50.5} & \textbf{50.8} \\
        \midrule
        \textbf{\textcolor{darkgreen}{$\Delta$ vs. SOTA}} & \textbf{\textcolor{darkgreen}{+3.0}} & \textbf{\textcolor{darkgreen}{+2.4}} & \textbf{\textcolor{darkgreen}{+1.7}} & -1.0 & \textbf{\textcolor{darkgreen}{+3.7}} & \textbf{\textcolor{darkgreen}{+3.3}} & \textbf{\textcolor{darkgreen}{+5.5}} & \textbf{\textcolor{darkgreen}{+1.5}} & \textbf{\textcolor{darkgreen}{+2.1}} & \textbf{\textcolor{darkgreen}{+0.2}} & \textbf{\textcolor{darkgreen}{+3.1}} &
        \textbf{\textcolor{darkgreen}{+4.2}} &
        \textbf{\textcolor{darkgreen}{+5.7}} &
        \textbf{\textcolor{darkgreen}{+5.6}}&
        \textbf{\textcolor{darkgreen}{+5.8}}\\
        \bottomrule
    \end{tabular}
    }
    \label{tab:main-results}
    \vspace{-2mm}
\end{table*}

\begin{table*}[!th]
  \centering
  \caption{\textbf{Zero-Shot Unsupervised Video Instance Segmentation.} By scaling up to 13K training videos, we show our method can scale to a powerful zero-shot model: Training on 13K real videos outperforms the model using 1.3 million synthetic videos.}
  \vspace{-2mm}
  \resizebox{\textwidth}{!}{
  \begin{tabular}{lccccccccccc}
    \toprule
    Dataset & & \multicolumn{2}{c}{YouTube-VIS 2019} & \multicolumn{2}{c}{YouTube-VIS 2021} & \multicolumn{2}{c}{YouTube-VIS 2022} & \multicolumn{2}{c}{OVIS} & \multicolumn{2}{c}{Cityscapes} \\
    \cmidrule(lr){3-4} \cmidrule(lr){5-6} \cmidrule(lr){7-8} \cmidrule(lr){9-10} \cmidrule(lr){11-12}
    Method & Training Data & AP$_{50}$ & AP & AP$_{50}$ & AP & AP$_{50}$ & AP & AP$_{50}$ & AP & AP$_{50}$ & AP \\
    \midrule
    VideoCutLER~\cite{wang2024videocutler} & 1.3M Syn. Videos & 48.2 & 24.5 & 37.0 & 18.0 & 31.7 & 15.0 & 4.2 & 1.1 & 1.7 & 0.9 \\
    S2D (Ours) & 13K Real Videos & \textbf{51.4} & \textbf{25.6} & \textbf{41.8} & \textbf{19.8} & \textbf{35.3} & \textbf{16.4} & \textbf{5.3} & \textbf{1.7} & \textbf{3.9} & \textbf{1.4} \\
    \midrule
    \textbf{\textcolor{darkgreen}{$\Delta$ vs. SOTA}} & & \textbf{\textcolor{darkgreen}{+3.2}} & \textbf{\textcolor{darkgreen}{+1.1}} & \textbf{\textcolor{darkgreen}{+4.8}} & \textbf{\textcolor{darkgreen}{+1.8}} & \textbf{\textcolor{darkgreen}{+3.6}} & \textbf{\textcolor{darkgreen}{+1.4}} & \textbf{\textcolor{darkgreen}{+1.1}} & \textbf{\textcolor{darkgreen}{+0.6}} & \textbf{\textcolor{darkgreen}{+2.2}} & \textbf{\textcolor{darkgreen}{+0.5}} \\
    \bottomrule
  \end{tabular}
  }
  \label{tab:zero-shot}
  \vspace{-2mm}
\end{table*}

\section{Experiments}
\label{sec:formatting}

\noindent\textbf{Training Datasets.} 
We perform in-domain experiments on YouTube-VIS (YTVIS) 2019 \cite{Yang2019ytvis} \& 2021 \cite{ytvis2021} as well as DAVIS-all~\cite{pontdavis2017}. For the YouTube-VIS datasets, we train on the training split with our Dual-Stage Distillation Training. For DAVIS-all, a traditional training-validation split does not exist. Hence, we train and evaluate only the stage 1 Sparse-To-Dense model on the dataset to demonstrate its ability to implicitly propagate instance masks. 
The previous state-of-the-art model, VideoCutLER~\cite{wang2024videocutler}, trains on 1 million synthetic videos from ImageNet. The dataset scale of their approach allows the model to generalize well, showing strong zero-shot performance. 
Hence, we investigate how our approach scales with a large video dataset, comprising video datasets outside of the domain of the video instance segmentation evaluation. For this, we combine the training splits of VIPSeg~\cite{miao2022vipseg}, MOSEv1~\cite{ding2023mose}, and all horizontal videos from SA-V~\cite{ravi2024sam2}. After the Keymask Discovery has filtered out unsuitable videos, the resulting dataset has a combined size of 13,000 videos, far surpassing YouTube-VIS which contains only 2,900 training videos. By training on this dataset with Sparse-To-Dense Distillation, we obtain a zero-shot model.

\noindent\textbf{Experiment Setup.} 
We first predict single-frame masks with CutS3D~\cite{sick2025cuts3d} on the respective training datasets. Next, we extract temporally-coherent keymasks and train our stage 1 Sparse-to-Dense distillation model on the sparse annotations. As mentioned for the DAVIS dataset, we only perform this stage to convert the sparse annotations to dense predictions. For the YouTube-VIS datasets, we perform both rounds of distillation training, as described in \cref{sec:dual-stage}. We apply the same weight initialization as Wang et al.~\cite{wang2024videocutler} and use a VideoMask2Former~\cite{cheng2021mask2formervideoinstancesegmentation} with a ResNet-50~\cite{he2016resnet} backbone, trained with video snippets comprised of 2 frames, as is standard for this model. After training, we keep the teacher model for evaluation. 

\noindent\textbf{Evaluation Datasets \& Protocols.} 
We evaluate our in-domain models for unsupervised video instance segmentation on the validation splits of YTVIS 2019 \& 2021~\cite{Yang2019ytvis, ytvis2021}, as well as DAVIS-all~\cite{pontdavis2017}. Furthermore, we evaluate semi-supervised and fully-supervised video instance segmentation. Here, the unsupervised model serves as a pre-trained weight initialization. We finetune the model on different percentages of labeled videos for semi-supervised learning, and the full dataset for supervised learning. Again, all models are evaluated on the YTVIS 2021 validation split.
In addition to YTVIS 2019 \& 2021, we evaluate our zero-shot model on YTVIS 2022 validation, OVIS~\cite{qi2022ovis} \& Cityscapes~\cite{cordts2016cityscapes}. For the latter, we extract instance annotations from the video panoptic segmentation Cityscapes-VPS dataset~\cite{kim2020vps}. For OVIS, we evaluate on the training split. We provide further details in the supplementary.

\subsection{Unsupervised Video Instance Segmentation}
\noindent\textbf{In-Domain Evaluation.}
After training our model in-domain, we report our results in \cref{tab:main-results}. Across all datasets, we compare our approach against baselines from VideoCutLER~\cite{wang2024videocutler}. This includes the image-trained CutLER~\cite{wang2023cut} model, and two approaches utilizing model-predicted optical flow estimations, OCLR~\cite{xie2022segmenting} and MotionGroup~\cite{yang2021motiongroup}.
For both YTVIS datasets, our model outperforms the best-performing baseline by up to \textbf{+2.1} AP. We show qualitative results on YTVIS-2021 in \cref{fig:qualitative}.
When evaluating the model's predictions on DAVIS-all, we find it outperforms VideoCutLER by \textbf{+5.7} $\mathcal{J}\&\mathcal{F}$. This improvement is despite only performing Sparse-To-Dense Distillation on the keymasks from DAVIS to obtain a full dense prediction set. \newline

\noindent\textbf{Zero-Shot Evaluation.} 
Beyond training and evaluating our model on in-domain data, we experiment with training a zero-shot model by learning exclusively from real videos. We start by constructing a large and diverse video dataset. We combine videos from VIPSeg~\cite{miao2022vipseg}, MOSEv1~\cite{ding2023mose} and SA-V~\cite{ravi2024sam2} and extract keymasks on this mixture-of-dataset. After this filtering, we obtain a dataset of 13,000 training videos with sparse pseudo-annotations. We perform Sparse-To-Dense Distillation on this dataset and find that one round is sufficient. 
After evaluating the general video instance segmentation capabilities of our model, we report zero-shot results in \cref{tab:zero-shot}. 
Across all evaluated datasets, our approach outperforms VideoCutLER by up to \textbf{+4.8} AP$_{50}$. This clearly emphasizes the value real-world data brings to the learning process. For Cityscapes, we find both models struggle. We assume their training data is to far out of domain for the driving scenarios.  

\subsection{Semi-Supervised \& Fully-Supervised \\ Video Instance Segmentation}
We experiment with semi-supervised \& supervised finetuning, using our models as pre-trained weight initialization.

\noindent\textbf{Evaluation Protocol.}
We re-use the weights of our model as unsupervised weight initialization, pre-trained on YTVIS-2021.
For the semi-supervised evaluation, we select varying amounts of annotated videos as labeled subset to finetune the model. For the fully-supervised evaluation, we train the model on the entire training dataset. We provide further training details in the supplementary.

\noindent\textbf{Results.}
We report our evaluation results on the YouTube-VIS 2021 validation split in \cref{tab:semisupervised}. For low-data regimes, such as using only the 10\% of labeled training data, our model outperforms VideoCutLER~\cite{wang2024videocutler} by \textbf{+2.7} AP$_{50}$.
For fully-supervised training, our model is competitive with the baseline. This is despite VideoCutLER being pre-trained on over a million synthetic videos, while we pre-train our model only on 13,000 real videos. 

\begin{table}[!th]
    \centering
    \footnotesize
    \caption{\textbf{Semi-Supervised \& Fully-Supervised Video Instance Segmentation.} We show that in low-data regimes, our model trained on real videos outperforms VideoCutLER with ImageNet training across various data splits on YouTubeVIS-2021~\cite{ytvis2021}. 
    }
    \vspace{-2mm}
    \resizebox{\linewidth}{!}{
    \begin{tabular}{rlccc}
        \toprule
        && \multicolumn{3}{c}{YouTube-VIS 2021} \\
        \cmidrule(lr){3-5}
        Pct. &
        & DINO\textsuperscript{*}\cite{caron2021emerging} & VideoCutLER\textsuperscript{*}\cite{wang2024videocutler} & S2D (Ours) \\
        \midrule
        \multirow{2}{*}{\textbf{1\%}} &
        AP      & 11.1 & 14.5 & \textbf{15.2} \\
        & AP$_{50}$ & 20.0 & 23.9 & \textbf{25.1} \\
        \midrule
        \multirow{2}{*}{\textbf{5\%}} &
        AP      & 9.4  & 14.3 & \textbf{15.4} \\
        & AP$_{50}$ & 18.9 & 22.8 & \textbf{25.4} \\
        \midrule
        \multirow{2}{*}{\textbf{10\%}} &
        AP      & 11.5 & 16.2 & \textbf{17.4} \\
        & AP$_{50}$ & 21.8 & 25.7 & \textbf{28.4} \\
        \midrule
        \multirow{2}{*}{\textbf{25\%}} &
        AP      & 12.3 & 20.9 & \textbf{21.1} \\
        & AP$_{50}$ & 20.9 & 32.8 & \textbf{33.1} \\
        \midrule
        \multirow{2}{*}{\textbf{100\%}} &
        AP      & 26.3 & 32.4 & \textbf{33.0} \\
        & AP$_{50}$ & 43.3 & 51.0 & \textbf{53.1} \\
        \bottomrule
    \end{tabular}}
    \label{tab:semisupervised}
    \vspace{-4mm}
\end{table}
\section{Ablations}
\label{sec:formatting}

We present ablations for the various steps of our method, starting with the components of our Keymask Discovery (\cref{tab:abl-keymasks}), then ablating the components of our model trainings (\cref{tab:abl-modeltrainings}) and finally comparing our implicit mask propagation to using a separate propagation algorithm (\cref{tab:abl-propagation}). Further ablations are presented in the supplement.

\noindent\textbf{Keymask Discovery Components.}
In \cref{tab:abl-keymasks}, we ablate all components of the Keymask Discovery. As a baseline, we train an Image Mask2Former~\cite{cheng2022mask2former} on the single-frame masks. We then compare training on video annotations from just visibility grouping to the sparse keymasks with proxy propagate-and-match. Finally, we compare to the model performance on the final dense labels. We report results on YouTube-VIS 2021, by training our video instance segmentation model on the resulting annotations in the dual-stage setup.
While we find that training a model on single-frame masks results in performance competitive with VideoCutLER~\cite{wang2024videocutler}, the model significantly improves with temporally-coherent sparse annotations from our keymask discovery. Furthermore, training on the final dense annotations, obtained through Sparse-To-Dense distillation, further improves the results.


\noindent\textbf{Model Training.} 
We ablate the modifications to the VideoMask2Former. As a baseline, we train the model on the pseudo-annotations without any modifications. We experiment with just adding pseudo-dense data loading, then we add the Temporal DropLoss. Finally, we compare against training the final model with dual-stage training. We again evaluate on YouTube-VIS 2021 and report the performance of our final dual-stage model training.
We see improvements of our method from implicit mask propagation with our Temporal DropLoss. When employing Sparse-To-Dense distillation in addition, our final model further improves to 20.1 AP.

\noindent\textbf{DINO Propagation vs. Sparse-To-Dense Distillation.} 
The purpose of Sparse-To-Dense Distillation and the Temporal DropLoss is to enable implicit mask propagation using a Video Instance Segmentation model. However, alternatives exist, such as explicit propagation algorithms~\cite{jabri2020crw, caron2021emerging}, i.e., propagating masks throughout a video by connecting semantic encodings of individual frames.
Hence, we compare using an off-the-shelf propagation algorithm based on DINO~\cite{caron2021emerging} semantics to our Sparse-To-dense Distillation on DAVIS-all~\cite{pontdavis2017}.
Since these algorithms are designed to propagate instance masks from a starting frame, we perform this propagation for all single-frame masks in the video, and select the best result per instance to report our results in \cref{tab:abl-propagation}. We find that our approach outperforms this propagation alternative.

\begin{table}[!ht]
  \centering
  \caption{\textbf{Keymask Discovery Components.} Metrics on YTVIS-2021. \textsuperscript{\textdagger}Single-Frame Masks trained with an Image Mask2Former, the weights are compatible with the VideoMask2Former.}
  \vspace{-2mm}
  \begin{tabular}{lcc}
    \toprule
    Components & AP$^{\text{mask}}_{50}$ & AP$^{\text{mask}}$  \\
    \midrule
    Single-Frame Masks\textsuperscript{\textdagger} & 36.5 & 17.8  \\
    \hspace{1.5mm} + Visibility Grouping & 38.9 & 18.9  \\
    \hspace{1.5mm} + Temporal Matching & 40.9 & 19.5 \\
    \hspace{1.5mm} + Dense Labelset & \textbf{42.5} & \textbf{20.1} \\
    \bottomrule
  \end{tabular}
  \label{tab:abl-keymasks}
  \vspace{-2mm}
\end{table}

\begin{table}[!ht]
  \centering
  \caption{\textbf{Model Design Choices.} We ablate the effect of modifications to our training pipeline on YTVIS-2021.}
  \vspace{-2mm}
  \begin{tabular}{lcc}
    \toprule
    Components & AP$^{\text{mask}}_{50}$ & AP$^{\text{mask}}$  \\
    \midrule
    VideoMask2Former & 31.2 & 14.0  \\
    \hspace{1.5mm} + Pseudo-Dense Dataloading & 39.9 & 19.3  \\
    \hspace{1.5mm} + Temporal DropLoss & 41.9 & 19.8 \\
    \hspace{1.5mm} + Sparse-To-Dense Distillation & \textbf{42.5} & \textbf{20.1} \\
    \bottomrule
  \end{tabular}
  \label{tab:abl-modeltrainings}
  \vspace{-2mm}
\end{table}

\begin{table}[ht]
    \centering
    \caption{\textbf{Sparse-To-Dense Distillation vs. DINO Propagation.} On DAVIS, we propagate every mask, then match with the ground-truth to calculate the J\&F.}
  \vspace{-2mm}
    \resizebox{\linewidth}{!}{
    \begin{tabular}{lccc}
    \toprule
    Method &  $J\&F$ & $J$-Mean & $F$-Mean \\
    \midrule
    Propagation with DINO~\cite{caron2021dino} & 42.2 & 43.4 & 41.0 \\
    Sparse-To-Dense Distillation & \textbf{50.6} & \textbf{50.5} & \textbf{50.8} \\
    \bottomrule
    \end{tabular}
    }
    \label{tab:abl-propagation}
  \vspace{-2mm}
\end{table}



\section{Discussion} In our experiments, we find that our approach struggles with segmenting smaller objects, as does the previous state-of-the-art. A potential solution could be implementing copy-paste augmentation for videos and scale down copied objects. 
Furthermore, our zero-shot model struggles for driving-domain data. We believe adding such domain-specific data to our dataset-mix can help the model improve.
While our method relies on deep motion cues to establish temporal coherence, future methods could improve this aspect by using deep temporal features.

\section{Summary}
In this paper, we have introduced S2D, a novel approach to learn unsupervised video instance segmentation from real video data. We have proposed a novel Keymask Discovery algorithm for identifying keymasks with temporal coherence from unsupervised image instance masks. To perform mask propagation for these sparse annotations, we have introduced Sparse-To-Dense Distillation with Temporal DropLoss. Finally, we have trained a self-distilled video instance segmentation model on the resulting dense annotations. Our final model outperforms the current state-of-the-art for unsupervised video instance segmentation, for both in-domain and zero-shot settings. 


{
    \small
    \bibliographystyle{ieeenat_fullname}
    \bibliography{main}
}



\end{document}